# Dynamic Move Chains – a Forward Pruning Approach to Tree Search in Computer Chess

Kieran Greer, Distributed Computing Systems, Belfast, UK.
http://distributedcomputingsystems.co.uk
Version 1.4

*Abstract*—This paper[1] proposes a new mechanism for pruning a search game-tree in computer chess. The algorithm stores and then reuses chains or sequences of moves, built up from previous searches. These move sequences have a built-in forward-pruning mechanism that can radically reduce the search space. A typical search process might retrieve a move from a Transposition Table, where the decision of what move to retrieve would be based on the position itself. This algorithm stores move sequences based on what previous sequences were better, or caused cutoffs. This is therefore position independent and so it could also be useful in games with imperfect information or uncertainty, where the whole situation is not known at any one time. Over a small set of tests, the algorithm was shown to clearly out-perform Transposition Tables, both in terms of search reduction and game-play results.

## 1   Introduction

This paper describes a new way of dynamically linking moves into sequences that can be used to optimise a search process. The context is to optimise the search process for the game of computer chess. Move sequences are returned during the searching of the chess game-tree that cause a cutoff, or determine that certain parts of the tree do not need to be searched. These move sequences are usually stored in Transposition Tables [11][12], but instead, they can be stored in a dynamic linking structure and used in other parts of the search process to reduce the search tree size. They can be used in the same way as Transposition Table entries by returning an already searched sequence of moves, which removes the need to search the tree structure that would have resulted in this move

---

[1] For an improved version of this paper, including new results and conclusions, please read http://www.hindawi.com/journals/aai/2013/357068/.

Also, please cite the paper as:
Kieran Greer, "Tree Pruning for New Search Techniques in Computer Games," Advances in Artificial Intelligence, Vol. 2013, Article ID 357068, 9 pages, 2013. doi:10.1155/2013/357068, Hindawi.

sequence. The term 'chain' instead of 'sequence' will be used to describe the new structure specifically. Move sequence is a more general term that can be used to describe any search move sequence.

This research has been carried out using an existing computer chess game-playing program called Chessmaps. The Chessmaps heuristic [8] was created as part of a DPhil research project that was completed in 1998. The intention was to try and add some intelligence into a chess game-playing program. If the goal is to build the best possible chess program, then the experience-based approach has probably solved the problem already, as the best programs are now better or at least equal to the best human players. Computer chess can also be used however simply as a domain for testing AI-related algorithms. It is still an interesting platform for trying to mimic the human thought process, or add human-like intelligence to a game-playing program. Exactly this argument, along with some other points made in this paper, are also written or thought about in [3] and [9]. While chess provides too much variability for the whole game to be defined, it is still a small enough domain to make it possible to accurately evaluate different kinds of search and evaluation processes. It provides complete information about its environment, meaning that the evaluation functions can be reliable, and is not so complex that trying to encapsulate the process in a comprehensive manner is impossible.

The rest of the paper is structured as follows: Section 2 describes the original Chessmaps Heuristic that can be used to order moves as part of a search process. Section 3 describes the dynamic move-linking mechanism that is the new research of this paper. Section 4 describes some other AI-related chess game-playing programs. Section 5 describes some test results from tests performed using the new linking mechanism. Section 6 describes how the linking mechanism could lead to other types of research, while section 7 gives some conclusions on the work.

## 2    Chessmaps Heuristic

The Chessmaps Heuristic [8] uses a neural network as a positional evaluator for a search heuristic. The neural network is taught to recognise what areas of the board a piece should be moved to, based on what areas of the board each side controls. The piece move is actually calculated based on what squares it attacks, or influences, after it has moved, which therefore includes the long range influence any piece.

The neural network is trained on Grandmaster or Master games. The chess position is converted into a board that defines what squares each side controls and this is then used as the input to training the neural network. The move played in that position is converted into the squares that the move played influences. These are the squares that the moved piece attacks and these are used as the output that the neural network then tries to recognise. The theory behind this is that there is a definite relation between the squares that a player controls and the squares that the player moves his/her pieces to, or the areas the player then plays to, when formulating a plan. The neural network by itself proved not to be accurate enough, but as it required the control of the squares to be calculated, this allowed several other move types to be recognised. One division would be to split the moves into safe and unsafe. Unsafe moves would lead to loss of material on the square the piece was moved to, while safe moves would not. It was also possible to recognise capture, forced and forcing moves. Forced moves would be ones where the piece would have to move because it could be captured with a gain of material. Forcing moves were moves that forced the opponent to move a piece because it could then be captured with a gain of material. This resulted in moves being looked at as part of a search process in the following order:

1. Safe capture moves.
2. Safe forced moves.
3. Safe forcing moves.
4. Safe other moves.
5. Unsafe capture moves.
6. Unsafe forced moves.
7. Unsafe forcing moves.

8. Unsafe other moves.

The neural network was really only used to order the moves in the 'other' moves category, although this would still be a large majority of the moves. The research therefore resulted in a heuristic that was knowledge-based, but also still lightweight enough to be used as part of a brute-force alpha-beta (α-β) search. Test results showed that it would reduce the search by more than the History Heuristic [11], but because of its additional computational requirements, it would use more time to search for a move and would ultimately be inferior. The heuristic however proved difficult to optimize, for example, trying to create quick move generators through bitmap boards; and so the only way to reduce the search time would probably be to introduce more AI-related techniques into the program. This has led to the following new suggestion for dynamic move sequences.

## 3   Dynamic Move Sequences

Two of the most popular experience-based approaches to minimising the search process are the History Heuristic [11] and Transposition Tables [11][12]. Tests have shown that using combinations of heuristics can produce a search reduction that is close to the minimal tree for an α-β search. This means that it is almost possible to produce the same search reduction that a perfect move ordering would produce. Transposition Tables however replace part of the search process with a result that has already been stored and so in effect forward prunes parts of the search tree. This means that it could produce a search tree size that was less than the minimal value, because parts of the search tree can be removed without being searched. The History Heuristic is attractive because of its simplicity and also its compact form. It can represent all of the moves in a single 64-square board array. Transposition Tables can become very large in size, therefore requiring an indexing system to search over thousands of entries or more, to find the position that relates to the one currently being searched. The question arises could it not be possible to represent the Transposition Table information in a more compact form, if all of the moves can be represented in a 64-square board array? This might not be possible for Transposition Tables, as they need to index position descriptions themselves, but the dynamic move sequences will use this sort of structure.

The main structure for storing dynamic move sequences is a 64-square board array. During a normal α-β search, when a move sequence is found that causes a cutoff in the search process, it can be stored in this array by storing the first move in the array element representing the first move's 'move from' square. The entry then also stores all of the other move information, such as the 'move to' square or the piece type. Multiple entries for different pieces can then be stored individually for the same square. This first move then stores a link to the second move, which stores a link to the next move, and so on. Each move could also have a weight value associated with it that would be incremented or decremented, based on whether the move sequence is subsequently found to be useful or not. The current implementation however does not require this, as any stored moves are automatically used again.

There are differences and similarities between the move chains and Transposition Tables. One thing is that these move sequences can be removed as well as added. Another is that Transposition Tables retrieve a move sequence for the 'position' being currently searched, whereas the dynamic move chains do not consider the position but retrieve a possible move sequence for the 'move' currently being looked at. There is also however a sort of similarity here, where any move played will lead to a position that could be stored in a table, or any position evaluation retrieved from a table will result in a move that could be stored in a move chain. The difference then is that the position evaluation is exact and static, whereas the move path evaluation is dynamic and needs to be checked again before it can be used.

The argument for storing general information in experience-based heuristics is the fact that positions in a search tree vary only slightly from one ply to the next and so if some information is retrieved, it is likely to be reliable, even if it does not exactly match the current position. Dynamic move chains can also use this argument for storing move sequences. The positions that the move chains are used in (over a single game) should be closely related to each other, so if the move sequence is legal, it might be useful in a different position as well. Unreliable sequences then need to be removed when necessary and this is helped by constantly updating the information on the stored move chains. Figure 1 is a diagram showing the potential difference in the number of nodes searched using a standard α-β search or using dynamic move chains. Note that this is only illustrative and is not meant to be accurate. Table 1 gives more accurate numbers. The figure is intended to show simply that the standard search is much broader and exhaustive. It will prune more, from the first branch to the last, as the window of acceptable moves narrows. The new algorithm will produce much narrower and deeper searches on every branch. The dashed lines show where possibly forward pruning has taken place. With fewer moves being considered, this is prone to miss critical moves, however.

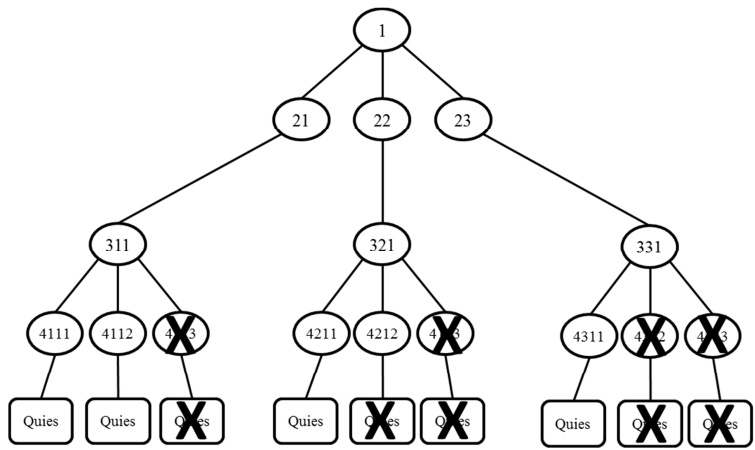

Figure 1a.

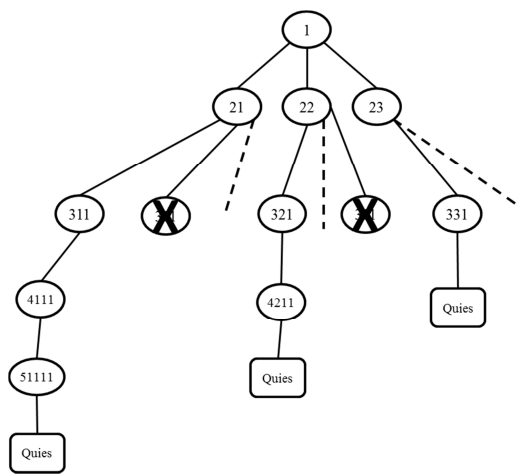

Figure 1b.

Figure 1. Diagram showing the difference in the number of nodes searched for a standard α-β search (1a) or a search with dynamic move chains (1b).

Figure 1a shows a game tree searched to a particular depth, with certain nodes being pruned through the α-β minimax algorithm. These nodes are crossed out with Xs. Considering the nodes in the subtree of the node labeled '21', for example, the diagram shows that there are 5 standard node evaluations and a further two quiescence searches after this. Note that a node must be evaluated before its sub-tree can be discarded. The quiescence search is much deeper, but considers only forcing or capture moves, until a quiescent or stable position is reached. If figure 1b represents the same tree, but including dynamic move chains, then the valid move sequences forward prune the tree by replacing large parts of it.  Only the paths in the move chains are considered and not the full sub-tree with multiple branches. This can occur at any level in the tree. It is also dynamic, where an incorrect evaluation will remove the chain and allow for a full search the next time the move is encountered.

## 4   Related Work

The brute-force programs are still the most successful. They rely more on program optimization and speed, to search more positions and therefore obtain a more accurate evaluation. There are also however a number of strong AI-related programs, some of which are listed here. A summary of recent advances in AI game-playing can be found in [5]. KnightCaps [1] is possibly one of the best AI attempts at playing computer chess, based on its playing strength. It uses a variation of the Temporal Difference machine learning algorithm called TDLeaf($\lambda$). The core idea of TD learning is that we adjust predictions to match other, more accurate, predictions about the feature. Accuracy can be obtained through deeper searches (more time), or evaluations of models that are closer to the known result. The same mechanism was also used in the NeuroChess program [12]. Blondie25 [4] is an example of another chess game-playing program that uses an evolutionary method to learn to play, as does [9]. These programs try to learn the evaluation function through feedback from playing a number of games, to adjust weights values related to features, to fine tune the evaluation criteria. They are therefore flexible in the sense that the criteria are not hard-coded beforehand and are improved through machine learning. In [3] they use a Bayesian network to adapt the program's game-playing strategy to its opponent, as well as to adapt the evaluation function.

As the tree search is already almost optimised, the most important aspect of the computer chess program is probably now the evaluation function. The optimisation results however are compared to the brute-force α-β search, but with forward pruning it is possible to produce even smaller search trees. The smaller the search tree, the more information that can be added to the evaluation function, and so trying to minimise the search further is still important. While not directly related to computer chess, the paper [2] describes recent advances in dynamic search processes. Their algorithm called Learning Depth-First Search (LDFS), aims to be both general and effective. It performs iterated depth-first searches enhanced with learning. They claim that LDFS show that the key idea underlying several existing heuristic-search algorithms over various settings is the use of iterated depth-first searches with memory for improving the heuristics over certain relevant states until they become optimal. Storing move chains is also a memory structure that is dynamic and built up over time. The dynamic moves approach probably also has similar intentions to the UCT search algorithms [7][14]. These also search to variable depths and select what branches to expand further. They are able to deal with large search trees or uncertainty in this way. The paper [7] describes how TD learning is offline, while UCT is online. It also states:

> 'At every time-step the UCT algorithm builds a search tree, estimating the value function at each state by Monte-Carlo simulation. After each simulated episode, the values in the tree are updated online and the simulation policy is improved with respect to the new values. These values represent local knowledge that is highly specialised to the current state.'

The move chains mechanism appears to be very different however. The saved moves do not rely on any kind of knowledge, apart from whether they are OK as part of the search process. In this research that means if they are legal move sequences and weighted values are not necessarily required to determine this. The bandit algorithms, such as UCT, appear to rely on aggregating statistical values, to build up a knowledgeable picture of what nodes are good or bad.

# 5   Testing

## 5.1   Search to fixed depth for different algorithm versions

This set of tests tried three different search algorithms on the same set of positions, but to fixed depths of 5 or 6 ply for each position. The search process automatically applies a minimal window and iterative deepening to each search that is performed. As well as this, the three algorithms tested included either the Chessmaps Heuristic by itself, the Chessmaps Heuristic and Transposition Tables, with 1000000 entries for each side; or the Chessmaps Heuristic with dynamic move sequences. Table 1 gives the results of the test searches.

|           | **Chessmaps** | **Chessmaps + TT** | **Chessmaps + MC** |
|---|---|---|---|
| **Depth 5**   | 653134.3918 | 438067.3505 | 15365.72165 |
| **Reduction** | 0%          | 33%         | 97.5%       |
| **Depth 6**   | 3263162.34  | 2467462.031 | 18920.76289 |
| **Reduction** | 0%          | 24.5%       | 99%         |

Table 1. Comparison of node count for 98 game positions with Chessmaps, plus either Transposition Tables or Move Chains.

The first line of numbers are the actual search results, while the second line is the percentage difference between Chessmaps by itself and the other mechanism that it is compared to. The reduction row describes the amount of extra search reduction that either the Transposition Tables or the move chains produce over just the Chessmaps heuristic. The Transposition Tables that are implemented as part of the Chessmaps program can be considered as a standard variation of the mechanism . They produce a significant reduction in the number of nodes searched, as shown in the table. The move chains then produce an even larger search reduction by radically forward pruning the search tree. Most of the generated move chains were only to depth 1, but could be to depth 3 or 4. The algorithm of Figure 2 describes how the move chains are used as part of a search process.

```
moveList = generateMoves();

while (moveList not empty) {
        nextMove = getNextMove();

        if (nextMove in moveChains) {
                moveSequence = getMoveSequence(nextMove, moveChains);
                if (isValid(moveSequence)) {
                        linkPath = moveSequence;
                        nextEval = eval( linkPath then quies search);
                } else {
                        remove moveSquence from moveChains;
                        linkPath = null;
                }
        } else {
                linkPath = null;
        }

        if (linkPath == null) {
                nextEval = eval(nextMove then full search + quies search);
        }

        if (nextEval > bestEval) {
                bestEval = nextEval;
                bestMove = nextMove;
        }

        if (bestEval > alpha) {
                alpha = bestEval;
                bestPath = serchPath;
                if (linkPath != null) update linkPath in moveChains;
        }
}

always add final bestPath to moveChains;
```

Figure 2. Search algorithm with move chains, highlighting where move chains are used.

This is probably just one variation of what could be quite a flexible algorithm and outlines where the move chains are used only. In this version, the move chains replace almost exactly where the transposition table would be used. Any entry is automatically used if legal, and any new entry automatically overwrites an existing entry, as it is considered to be closer to the final solution. Any entry found not to be legal is automatically removed. The traditional assumption is that this should invalidate the search process. The results of playing the two algorithms against each other, as described in section B however, shows that this is not the case. If considering only the move chains that are generated for a search to depth 5: a quiescence search could search 15 to 40 times more nodes than the α-β search. Dynamic move chains would be retrieved only in the α-β search and for possibly ⅓ to ½ of the nodes searched, but this is comparing moves to nodes. It is estimated that on average, 6 moves are evaluated in a middlegame position. The chain would be used possibly ½ to ¾ of the time a chain was retrieved. So can an argument be made for why the shallow move chain searches might also work? Suppose that a move chain is stored that is only one move deep. This is only stored if the move was evaluated as better or caused a cutoff, which is only after a stabilizing quiescence search as well. Also, consider the fact that only the first positions in a search process are searched to the full depth. The leaf nodes are only searched to a depth of 1, plus the extensions. So this algorithm is treating more of the nodes like this. Because it is dynamic, the evaluations of the stored moves are constantly changing and being updated, which will help to maintain its accuracy. The previous searches compensate for the shallow depth of a move chain.

**5.2   Playing Different Algorithms Against Each Other**

The computer program is able to play different algorithms against each other. In these tests, the Chessmaps algorithm with Transposition Tables was played against the Chessmaps algorithm with the dynamic move chains. Both versions used an iterative deepening search with minimum window as well. There is also a random opening selector, so each game started from a different position. Tests were run over 100 games - 50 with Chessmaps plus Transposition Tables as White and 50 with Chessmaps plus move chains as White. Each side was allowed 30 minutes to play each game, with the results shown in Table 2. These results are not meant to be definitive, but should show that the new search algorithm is a viable alternative, with the idea of using dynamic move chains being a sound one. These results also show that using the dynamic move sequences with the Chessmaps heuristic outperforms using Transposition Tables with the Chessmaps heuristic, with the move chains version winning more games.

| Draw | MinWin+CM+TT | MinWin+CM+MC |
|---|---|---|
| 37 | 26 | 37 |

Table 2. Results of one hundred 30 minute games between Transposition Tables (TT) versus Move Chains (MC).

## 6  Future Work

The dynamic move chains are a bit like dynamic chunking, or tactical chunks, where a player would recognise and play combinations he/she has seen before. Transposition Tables are not 100% accurate because the hash function maps different positions to each entry, so there is some generalisation there as well. Therefore in this case, the table stores move sequences instead of positions that can maybe be thought of as tactical chunks. Why compare to a tactical chunk? The aim of these chunks is to suggest relevant moves in a position. Tactics are about moves, not positional evaluation and so move sequences known to be good could be thought of as chunks of tactical knowledge. It is also worth noting that many games now are games with imperfect information, meaning that all of the information is not always available. With perfect information, such as for chess, the whole board position is always known. With imperfect information there is missing information, which means that it is not possible to completely evaluate the whole situation, but some level of guessing is required. This would mean that if the move sequences could be considered independently, without requiring the whole board position, it could be useful in imperfect games as well.

Previous work on knowledge-based approaches to playing computer chess tried to recognise key features in a chess position that could be used to determine what the best move might be. It is argued that stronger players are able to recognise these features, or chunks of knowledge [6], and use them better to analyse a position and determine what the best move might be. One knowledge-based approach to building a computer chess program was to ask an expert to define these features in many different positions, along with the best moves in those positions. If this was a manual process then it suffered from problems such as the lack of generality, with the sheer number of positions that would need to be analysed and defined to make the process worthwhile. There would also be a problem of creating a language that could be used to successfully define these key features, or chunks, in a form that a computer could understand and then generalise over (the knowledge-acquisition bottleneck).

One future research direction would be to store the positions that each move sequence was played in with the move sequence itself. The set of positions can then be analysed to try and determine what the key features are. This is similar to the chunks of knowledge written about previously, but the process is slightly automated and there is some help in determining what to analyse. For one thing, the move sequences are already defined. They are determined by the feedback from the search process. The positions that the moves were played in are also defined and so the problem is now to recognise the key difference or similarities in the related positions and not to re-define these key features from scratch. This is still very challenging however and it is not clear how it might be done. Figure 3 describes this sort of process again.

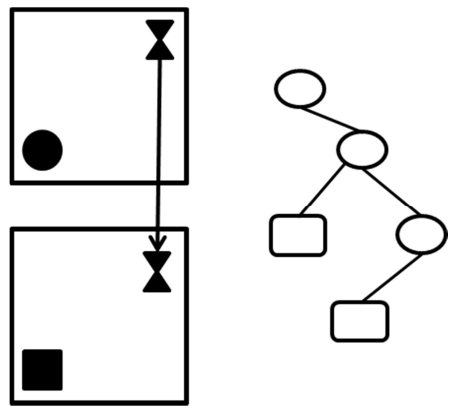

Figure 3. Move chain related to two different positions. The similarity in the positions is shown by the hour-glass feature.

In Figure 3, there is a dynamic move chain and also the two positions that it was played in. These positions share one common feature shown by the hour-glass object. If this similarity can be automatically recognised, then the process of storing dynamic move chains gives an automatic process both for recognizing what the critical move sequences in a game are and also what the critical features that allowed the moves to be played are.

# 7   Conclusions

These tests are based on a relatively small number of examples, but they show the potential of the algorithm and confirm that dynamic move chains are at least reliable. That is, they can be added to a search process and used without returning unreliable moves. Playing against the algorithm also confirms that it is not likely to make a poor move any more often than any of the other algorithms. So in this respect, move sequences or chains are a viable alternative to Transposition Tables. They also offer a compact solution to indexing the moves and offer new opportunities for research into other more knowledge-based or intelligent ways of playing the game of computer chess. The main danger is probably simply the loss in evaluation quality because of the reduction in the search. Most of the generated move chains were only to depth 1, but could be to depth 3 or 4, for example. So it is the ability to dynamically update them, the similarity between positions in a search and the quiescence search that keeps the search result accurate.

The test results suggest that the potential of the move chains might be the fact that it can provide a basis for new ways to search a game tree and even under different circumstances. It is shown to be at least equal to the other variants tested, but the different way that it stores information could lead to different search mechanisms that might ultimately prove better in certain types of game. It is not completely clear if this process can be mapped to a human thought process when playing the game. Re-trying a combination that a player thought of previously usually requires some positional judgment first. So it might simply be the statistical argument that supports the other experience-based heuristics. The position changes only slightly from one ply to the next and so if a move is found to be good, then there is a reasonable chance that it will be good in a future position as well. The intention of this research has been to reduce the search process by so much that the evaluation function can be made more substantial, to make the whole game-playing process more human-like. In this respect, the research has certainly achieved its goal.